\documentclass{article}

\PassOptionsToPackage{numbers, compress}{natbib}

\usepackage[preprint]{neurips_2024}

\usepackage[utf8]{inputenc} 
\usepackage[T1]{fontenc}    
\usepackage{hyperref}       
\usepackage{url}            
\usepackage{booktabs}       
\usepackage{amsfonts}       
\usepackage{nicefrac}       
\usepackage{microtype}      
\usepackage{xcolor}         
\usepackage{graphicx}
\usepackage{amsmath}
\usepackage{wrapfig}
\usepackage{subcaption}
\usepackage{enumitem}
\usepackage{amsmath,amssymb,amsfonts}

\title{Pathway-Guided Optimization of Deep Generative Molecular Design Models for Cancer Therapy}

\author{%
Alif Bin Abdul Qayyum$^{1}$\thanks{These authors contributed equally to this work.} \quad Susan D. Mertins$^{2}$\footnotemark[1] \quad Amanda K. Paulson$^{3}$\footnotemark[1] \\
\quad \textbf{Nathan M. Urban}$^4$ \quad \textbf{Byung-Jun Yoon}$^{1,4}$ \\
$^1$Department of Electrical and Computer Engineering, 
Texas A\&M University, \\
College Station, TX 77843-3128, USA. \\
\quad $^2$Cancer Data Science Initiative, 
Frederick National Laboratory for Cancer Research, \\
Leidos Biomedical Research, \\
Frederick, MD 21702, USA. \\
\quad $^3$Department of Pharmaceutical Chemistry, University of California, \\
San Francisco, CA 94158, USA.\\
\quad $^4$ Computational Science Initiative, Brookhaven National Laboratory, \\
Upton, NY 11973, USA. \\
}

\begin{document}

\maketitle

\begin{abstract}
  The data-driven drug design problem can be formulated as an optimization task of a potentially expensive black-box objective function over a huge high-dimensional and structured molecular space. The junction tree variational autoencoder (JTVAE) has been shown to be an efficient generative model that can be used for suggesting legitimate novel drug-like small molecules with improved properties.  While the performance of the generative molecular design (GMD) scheme strongly depends on the initial training data, one can improve its sampling efficiency for suggesting better molecules with enhanced properties by optimizing the latent space. In this work, we propose how mechanistic models - such as pathway models described by differential equations - can be used for effective latent space optimization(LSO) of JTVAEs and other similar models for GMD. To demonstrate the potential of our proposed approach, we show how a pharmacodynamic model, assessing the therapeutic efficacy of a drug-like small molecule by predicting how it modulates a cancer pathway, can be incorporated for effective LSO of data-driven models for GMD.
\end{abstract}
\section{Introduction}
\label{sec:intro}
Cancer is the second leading cause of death in United States and a major public health concern across the globe \cite{whoCancer}, posing the highest clinical, social, and economic burden among all human diseases \cite{cancer-burden, economic-burden, social-burden}. \newline
Traditional chemotherapy has significant toxicity and side effects due to its inability to distinguish between tumor and normal tissue. This has led to a tremendous shift in cancer research, driven by interest among researchers in drug-like molecule design with high specificity and low off-target effects for cancer therapy \cite{small-molecule-cancer-therapy, targeted-cancer-therapy, small-molecule-cancer-therapy-2, small-molecule-cancer-therapy-3, small-molecule-cancer-therapy-4, cancer-therapy-compare-antibody-small-molecule}. Computation-driven generative molecular design that optimizes multiple drug-like properties at once can be formulated as a mathematical optimization problem of an objective function over a high-dimensional, structured and complicated input space. Optimizing drug-like molecule efficacy is a daunting endeavor for two significant reasons. First, it is difficult to search through millions or even billions of natural and artificial molecules to identify an optimal candidate for modulating the function of a protein that drives the specific disease in consideration, similar to the problem of finding a needle in a haystack \cite{small-molecule-design-challange}. Second, even if a hit is discovered, establishing a precise model to understand the connection between the protein's modulation and the effect on the disease in question poses a formidable challenge \cite{disease-and-protein}. \newline
Since the advancement in deep learning, researchers have pursued novel deep generative models to computationally generate new ideas. Some of the most used methods among deep generative models are variational autoencoders (VAE) \cite{original-vae, vae-book, vae-tutorial}, generative adversarial networks (GAN) \cite{original-gan, gan-intro-outlook, gan-review, gan-theory-applications}, diffusion models \cite{ddpm, improved-ddpm} and consistency models \cite{consistency}. Due to the presence of a continuous low-dimensional latent space, a VAE is most suitable in conditional data generative tasks which optimize an objective function over an input space with the main goal of generating new data points \cite{conditional-vae}. Many promising results have surfaced in such optimization problems, such as conditional image generation \cite{conditionalimagelatentspace, conditionalimagepixelcnn}, text generation \cite{text-gen}, molecular and materials design \cite{molecule-gen, molecule-gen-review}, and neural architecture search \cite{neural-search}. \newline
Even after the success of deep generative models in various research fields, optimization of a deep generative model over a specific and hard to evaluate objective function such as cancer therapeutic capability of drug-like molecules on scarce structured input data, e.g. molecular structures, is still an area of open research \cite{drug-design-machine-to-deep-learning}. Recently, latent space optimization (LSO) has shown very promising results in optimization tasks on high dimensional structured input data \cite{bombarelli-vae}. Additionally, in order to effectively optimize the latent space for molecular property enhancement, it is crucial to have a dependable evaluation function. A rule-based model eliminates the requirement for a labeled dataset, offering a novel approach to directly optimizing the generative model to produce molecular structures with superior cancer therapeutic scores. \newline
We introduce an innovative approach for optimizing deep molecular generative models through incorporation of a rule-based, pathway-guided mechanistic model to evaluate the therapeutic efficacy of generated molecules. To the best of our knowledge, this work represents the pioneering effort to integrate a mechanistic model as an evaluative component within the optimization framework of deep molecular generative models. The outline of this paper is as follows: we discuss some backgrounds of our work in Sec, \ref{sec:background}, explain our proposed methodology in Sec. \ref{method}, demonstrate the observed results in Sec. \ref{result} and discuss some limitations of our proposal in Sec. \ref{limitations}.
\section{Background}
\label{sec:background}
The inherent complexity of graph-like molecular structures, with their arbitrary connectivity, poses a challenging obstacle when attempting to arrange their construction into a sequential process. Furthermore, the discrete nature of decision-making involved in establishing the order for incremental graph construction renders such processes non-differentiable. \newline
Several different deep generative molecular design models have been proposed for generation of valid molecular structures. GraphVAE proposes an approach for generating a probabilistic fully-connected graph of a predefined maximum size directly at once, sidestepping the hurdles involved in incremental construction of molecular graphs \cite{graphvae}. Generative models like GraphVAE frequently encounter the issue of producing invalid data points when working with structured data, such as drug-like small molecules. Grammar variational autoencoder generates molecules by representing them as parse trees from a context-free grammar and ensures that the generated molecular structures are syntactically valid \cite{grammar-vae}. In order to maintain semantic validity when generating molecules, such as adhering to the SMILES language's requirement that generated rings must be closed, syntax-directed variational autoencoders enforce constraints on the output space so that the model not only generates syntactically valid molecules, but also the molecular structures are semantically reasonable \cite{sd-vae}. Structure generating variational autoencoders leverage Bayesian optimization to fine-tune a low-dimensional continuous latent space \cite{sg-vae}. GANs have also been used for molecule generation. MolGAN combines GAN with reinforcement learning to generate molecule structures with specific desired chemical properties \cite{molgan}. GraphGAN proposes an approach for learning the molecular graph connectivity distribution among atoms in a molecule \cite{graphgan}. Denoising diffusion probabilistic models have recently gained popularity for molecule generation task \cite{torsional-diffusion, geometric-diffusion, equivariant-diffusion-molecule, diffusion-prior-bridges}. \newline
A junction tree variational autoencoder (JTVAE) has been proposed to address the problem of molecular optimization by directly encoding and decoding molecular graphs instead of generating linear SMILES strings which can contribute to uncertainty and error\cite{jin2018junction, bombarelli-vae}. The autoencoder trains a deep neural network, consisting of an encoder-decoder pair, on hundreds of thousands of existing chemical structures. The encoder maps molecular structures into low-dimensional latent space and the decoder converts latent space vectors back into molecular structures, along with a predictor network predicting the property of molecular structures from the low-dimensional latent representations. To allow generation of valid chemical structures, the JTVAE initiates the process by generating a scaffold in a tree-like structure, which encompasses chemical substructures. It subsequently merges these substructures into a complete molecule using a graph message passing network, thereby enabling the gradual expansion of molecules while ensuring their chemical integrity is preserved at each stage of development. \newline
The optimization of a generative molecular design model, such as JTVAE, involves fine-tuning the model in a manner that encourages it to generate molecular structures that align with the desired optimal properties. In this process, we adjust the model's parameters, training procedures, or other relevant aspects to influence its output. The ultimate aim is to guide the generative model towards producing molecules that exhibit the specific properties we desire, which serves as our primary optimization objective. Latent space optimization (LSO) has shown promising results in optimization tasks on high dimensional structured input data, e.g. molecular structures \cite{bombarelli-vae}. LSO consists of two stages of computation. First, a latent space based deep generative model, such as VAE, is trained to encode tensors, representing structured input data, into a low-dimensional continuous space. This effectively converts the optimization problem of high dimensional structured input data into an optimization problem on low-dimensional continuous data. Finally, the objective function is fine-tuned to guide the decoder model in mapping the latent space to data points that exhibit optimized objective function values. LSO has found applications in diverse domains, including automatic machine learning \cite{auto-ml-1, vae-acyclic-graphs, sg-vae}, conditional image generation \cite{neural-architecture-optimization, conditionalimagelatentspace}, and enhancing model interpretability \cite{explain-uncertainty}. \newline
Sample efficient periodic weighted retraining combines the idea of latent space optimization with a JTVAE model \cite{NEURIPS2020_81e3225c}. This approach proposes weighting of the data distribution and periodic retraining of the generative model with Bayesian optimization on the latent space to generate molecules with specific desired properties. At each periodic retraining iteration, a set of sampled molecules from the latent space with desired properties is appended to the existing training dataset to persuade the generative model to map the latent space more into a desired direction to optimize an objective function. A similar periodic weighted retraining of generative model uses the idea of pareto-front optimality for multi-objective optimization \cite{mobo-nafiz}. \newline
In order to effectively optimize the latent space for molecular property enhancement, it is crucial to have a dependable evaluation function. A rule-based model eliminates the requirement for a labeled dataset, offering a novel approach to directly optimizing the generative model to produce molecular structures with superior cancer therapeutic scores, in contrast to an indirect approach of optimizing a basic protein inhibition constant (IC50). In this work, we incorporate periodic retraining as a strategy to improve the navigation through a lower-dimensional representation of the chemical space. This enhancement enables us to efficiently identify a subset of the chemical space that contains candidates with optimal molecular properties. The utilization of LSO serves to reduce the number of candidate molecules that must be virtually screened before reaching an optimal subspace. We demonstrate that LSO can flexibly optimize toward quantitative, QSAR-predicted properties like single protein inhibitory constants or more complex outcomes like therapeutic score as predicted by a cell-level mechanistic pathway model. In particular, we investigate the use of mechanistic models, such as rule-based pathway-guided mechanistic model described by differential equations, as a powerful approach for optimizing the latent space in the context of generative molecular design for cancer therapy. This is especially valuable in scenarios where data is limited, and traditional data-driven models may not be as effective. \newline
Pharmacodynamic models capture biochemical pathways that underscore cellular function and offer a means to describe drug-like small molecule downstream activity. In particular, ordinary differential equations, solved either deterministically or stochastically, are parameterized with measured values such reaction rates and species concentrations and lead to a kinetic solution \cite{braakman2022evaluation}. These mechanistic or pathway models have been utilized in the context of comprehending cancer carcinogenesis and possess the capacity to delineate a targeted therapeutic objective, such as inducing programmed cell death, also known as apoptosis. Other investigators have provided useful mechanistic models that predict drug timing \cite{rukhlenko2018dissecting} and combinations \cite{schmucker2021combination}, both important challenges in predictive oncology. As an example, we use the DNA damage response pathway and specifically focus on PARP1 inhibition. \newline
poly(ADP-ribose) polymerase 1 (PARP1) is a key protein that initiates the DNA damage response to single stranded breaks (SSBs) and other DNA lesions \cite{lord-and-ashworth}. PARP1 binds to DNA damage and PARylates histones and other protein targets, which in turn recruit downstream DNA repair enzymes. Proliferating tumor cells are under replicative stress and often have increased DNA damage, making PARP a promising target for cancer therapies. PARP inhibition induces synthetic lethality in BRCA mutant tumors and other tumors that exhibit defects in the double strand break (DSB) repair pathway \cite{Farmer}. When SSBs are not properly repaired, they can progress to DSBs which can’t be repaired by the mutant BRCA proteins, leading to eventual cell death. Several clinical PARP inhibitors are on the market that perform well for BRCA-mutant tumors, but they suffer from dose-limiting toxicities and therefore do not perform well as single agents \cite{Illuzzi}. Design of more potent inhibitors with less off-target effects and additional favorable ADME and medicinal chemistry properties represents an excellent use-case for multi-objective optimization.\newline
The primary objective of this research is to develop an optimization approach that harnesses the generative capabilities of JTVAE to generate novel, synthesizable molecular structures and to utilize a rule-based pathway-guided mechanistic model as the evaluation function for the cancer therapeutic efficacy property to be optimized. 
\section{Methods}
\label{method}
Let, $\mathbb{X}$ be the molecular structure input space and let $f: \mathbb{X} \mapsto \mathbb{R}$ be the objective function, the cancer therapeutic efficacy evaluation function of the molecular structure. $\mathbb{X}$ is high dimensional and structured, and $f(\mathbf{x})$ is a black-box, hard to evaluate function and not many points, $\mathbf{x} \in \mathbb{X}$ with objective function value $f(\mathbf{x})$ is available in hand. Let, $\mathbb{Z}$ be the latent space and $D(\mathbf{z}), \mathbf{z} \in \mathbb{Z}$ be the the generative model that transforms the latent space into molecular structures. The effective and efficient latent space based generative molecular design with an optimization objective problem seeks to optimize $f$ over $\mathbb{X}$ by training $D$, in a manner that $D$ transforms $\mathbb{Z}$ into $\mathbb{O} \in \mathbb{X}$ where $\mathbb{O}$ consists of molecular structures with optimal $f(\mathbf{o}), \mathbf{o} \in \mathbb{O}$.
\begin{wrapfigure}{r}{0.65\textwidth}
\centering
    \includegraphics[width=0.65\textwidth]{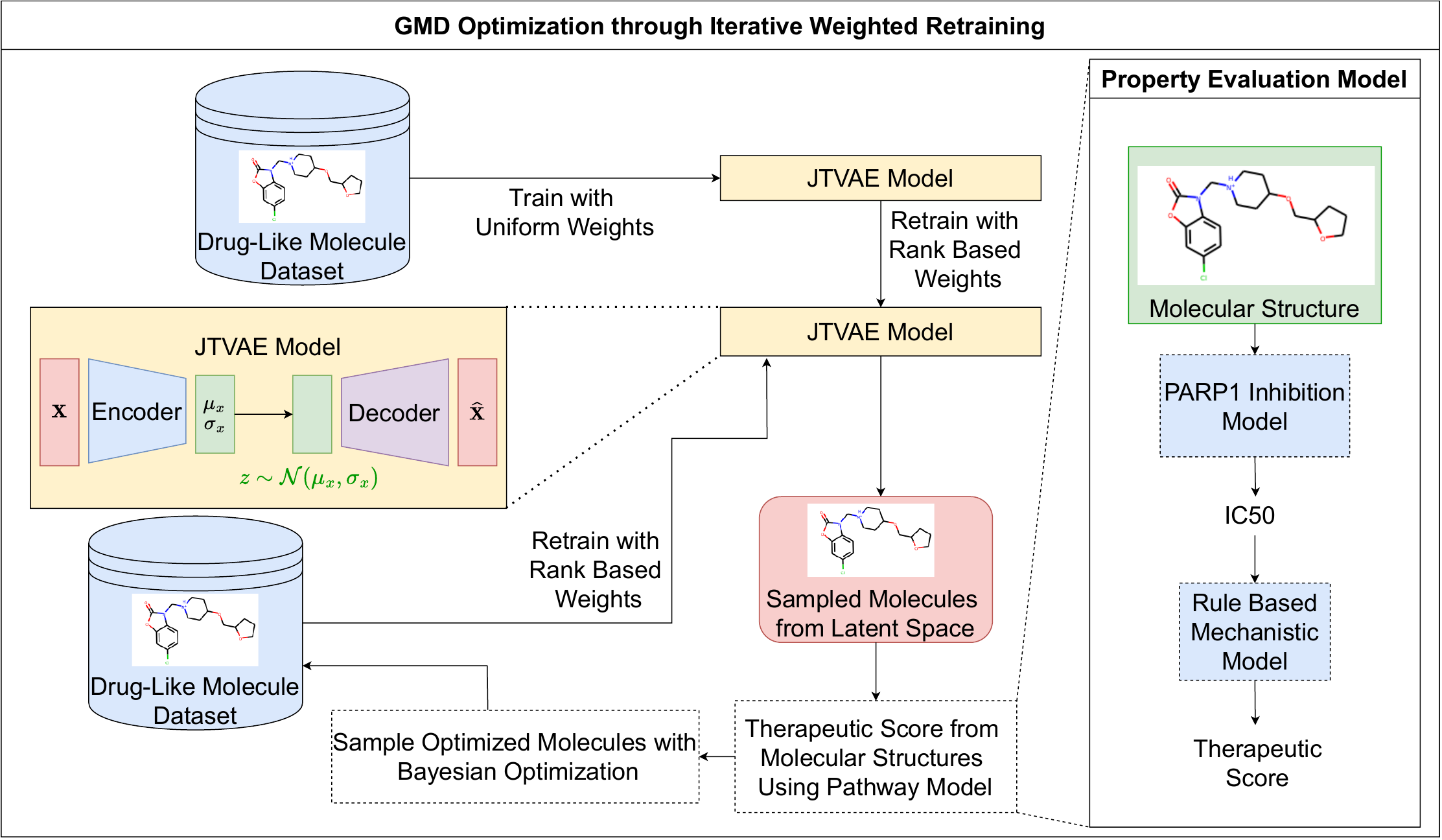}
    \caption{Overview of methodology}
    \label{workflow}
\end{wrapfigure}
The initial phase of the proposed method involves employing a JTVAE model, consisting an encoder $E: \mathbb{X} \mapsto \mathbb{Z}$ and a decoder $D: \mathbb{Z} \mapsto \mathbb{X}$, that has been trained with equal weights assigned to all molecules in the training set $\mathbb{D}$. Subsequently, we refine the model using a rank-based weighted retraining technique. In contrast to uniformly weighting all molecules, we allocate distinct weights to each molecule based on the specific property value we aim to enhance. 
To evaluate the cancer therapeutic score, we use a combination of data-driven and mechanistic model, consisting a machine learning based PARP1 inhibition model for IC50 evaluation and a rule-based pathway-guided mechanistic model for therapeutic score evaluation, as an approximation to $f$. We sample optimal points from the latent space of the JTVAE following Bayesian optimization process, which are used for subsequent refinement of the JTVAE model. The whole methodology can be best described by Fig. \ref{workflow}.
\subsection{Unweighted JTVAE training}
JTVAE represents an expansion of the latent space-driven deep generative model, known as a variational autoencoder, designed specifically for molecular graphs. JTVAE approaches the interpretation of molecules by considering them as assemblies of valid chemical components, thus avoiding the more intricate process of constructing molecules atom by atom. It accomplishes automated molecule design by following a two-step procedure. Firstly, it generates tree-like molecular substructures, breaking down the molecule into its constituent parts. Finally, it combines these substructures using a message passing graph neural network. This network allows for the integration of information between the substructures, ultimately resulting in the creation of molecules tailored to specific chemical properties \cite{jin2018junction}. A molecule is represented by two complimentary representations, the original molecular graph and its associated junction tree. The whole encoder $E$ consists of two distinct encoders $q(z_{T}|T)$ and $q(z_{G}|G)$ for encoded representations of the junction tree and graph structure accordingly. The latent representation $\{z_{T}, z_{G}\}$ is then decoded back to the molecular graph by first decoding the tree structure from its latent representation using the tree decoder, $p(T|z_T)$, then by decoding to the molecular graph from the decoded tree and latent graph using the graph decoder $p(G|z_G, T)$. Here, the tree decoder $p(T|z_T)$ and the graph decoder $p(G|z_G, T)$ consists the overall decoder $D$. During the optimization process, the loss function value for all molecules are aggregated by assigning weights $w_{i \in \{1, 2, \dots, |\mathbb{D}|\}}$ on each molecule in the training data $\mathbb{D}$. Within the unweighted training scheme, every molecule is treated with the same level of importance or significance during the training process. The weights for all molecules in the training data are equal, hence we refer it as unweighted JTVAE training. This indicates that no distinctions or preferences are made among the molecules regarding their individual contributions to the learning process. We utilize a JTVAE model that has previously undergone training, without any specific weighting scheme, to generate novel molecules, primarily focusing on different chemical properties such as logP or QED. In practical terms, this means that all molecules within the training set were assigned uniform weights, without any differentiation based on their individual attributes or characteristics. 
\subsection{Periodic weighted JTVAE training}
\label{weighted-train}
Our approach involves a synergy of techniques, namely data weighting and periodic retraining, aimed at elevating the generative model from a passive role as a decoder to an active contributor in the optimization process. This transformation is orchestrated to fine-tune the latent space, making it increasingly proficient at generating molecules with high property values over time.
\subsubsection{Data weighting}
\label{data-weight}
The quest for designing a generative model that is most conducive to optimization within the latent space remains an ongoing research endeavor. One effective strategy is to focus exclusively on the subset containing data points characterized by high property values. This approach is beneficial because it concentrates the modeling efforts on the most promising and desirable data instances, which can lead to more accurate and effective results. But with limited or scarce data in hand, it becomes advantageous to train the model using a probabilistic distribution that exhibits a high likelihood for data points with superior property scores while assigning a notably lower likelihood to those with lower property scores. This distribution essentially emphasizes the importance of high-scoring data points during the training process, enabling the model to make the most out of the limited data available and potentially enhance its performance on the desired property optimization task. We apply the later approach by assigning an explicit weight $w_i$ to each data point, such that $\sum_i w_i = 1$. We follow the rank-based weight function proposed in \cite{NEURIPS2020_81e3225c}. The weighting approach can be described by the following equations:
\begin{equation}
    rank_{f, \mathbb{D}}(\mathbf{x}) = | \{\mathbf{x}_i: f(\mathbf{x}_i) > f(\mathbf{x}), \mathbf{x}_i \in \mathbb{D}\}|
\end{equation}
\begin{equation}
    w(\mathbf{x}; \mathbb{D}, k) \propto \dfrac{1}{10^{-k}N + rank_{f, \mathbb{D}}(\mathbf{x})} 
\end{equation}
This weighting function assigns a weight roughly proportional to the reciprocal (zero-based) rank of each data point. Also, a high $k$ value assigns more weights on the high value points compared to the low value points.
\subsubsection{Periodic retraining}
To optimize the latent space of the JTVAE, periodically retraining the generative model is a simple but effective solution. In order to ensure the latent space remains effectively optimized \cite{active-learning} and to prevent the occurrence of catastrophic forgetting \cite{catastrophic-forgetting}, we employ a systematic approach of periodic retraining for the JTVAE model. This process involves several key steps, among which are weighted retraining, finding optimum latent samples from the latent space, appending those optimal molecules to the training set and iterative retrain of the JTVAE model. We periodically retrain the JTVAE model following the weighting scheme discussed in Sec. \ref{data-weight}, find optimum samples from the latent space, decode those latent samples back to molecular space, append those optimum samples to the training set, and then continue this iterative retraining process for a number of times. Rather than subjecting the JTVAE model to a complete retraining cycle using the entire updated training set, we adopt a more selective approach during each retraining iteration. We carefully choose only a portion of the training set for retraining the JTVAE model at each iteration. Also to maximize the optimization of the latent space, we ensure that the optimal molecules sampled during one retraining iteration are incorporated into subsequent iterations of retraining. In other words, the valuable insights gained from the selected optimum molecules in one iteration continue to influence and enhance the model to understand the latent space in subsequent iterations. By implementing this approach, we strike a balance between updating the model's knowledge with new data and preserving the knowledge gained from optimal molecules, resulting in a more effective and efficient optimization of the latent space over multiple iterations. To find the optimum latent samples, we follow Bayesian optimization. Bayesian optimization sustains a probabilistic model of the target function in order to select new points for assessment, guided by the modeled distribution of the objective values at points that have not yet been observed \cite{bo_tutorial, bo-tutorial-rl, bo-ml}. Our objective function is a rule-based pathway guided mechanistic model to predict the therapeutic score of a molecule. Our evaluation function consists of two parts: a machine learning model to predict the IC50 value from the molecular graph, and a rule based pathway model to predict therapeutic score from the predicted IC50 value. 
\subsection{Therapeutic Score Calculation}
The calculation of the therapeutic score is performed on two steps. First, the pIC50 value of the generated molecule is calculated using a machine learning based PARP1 inhibition model, which is then used to predict the IC50 values of new compounds. Then from the IC50 values we calculate the therapeutic score using an rule-based pathway-guided mechanistic model. We use three different versions of the rule-based pathway-guided mechanistic model. 
\begin{itemize}
    \item Physiologically viable pathway model.
    \item Modified pathway model.
    \item Physiologically impractical pathway model.
\end{itemize}

\begin{wrapfigure}{r}{0.5\textwidth} 
\centering
    \vspace{-6mm}
    \includegraphics[width = 0.5\textwidth]{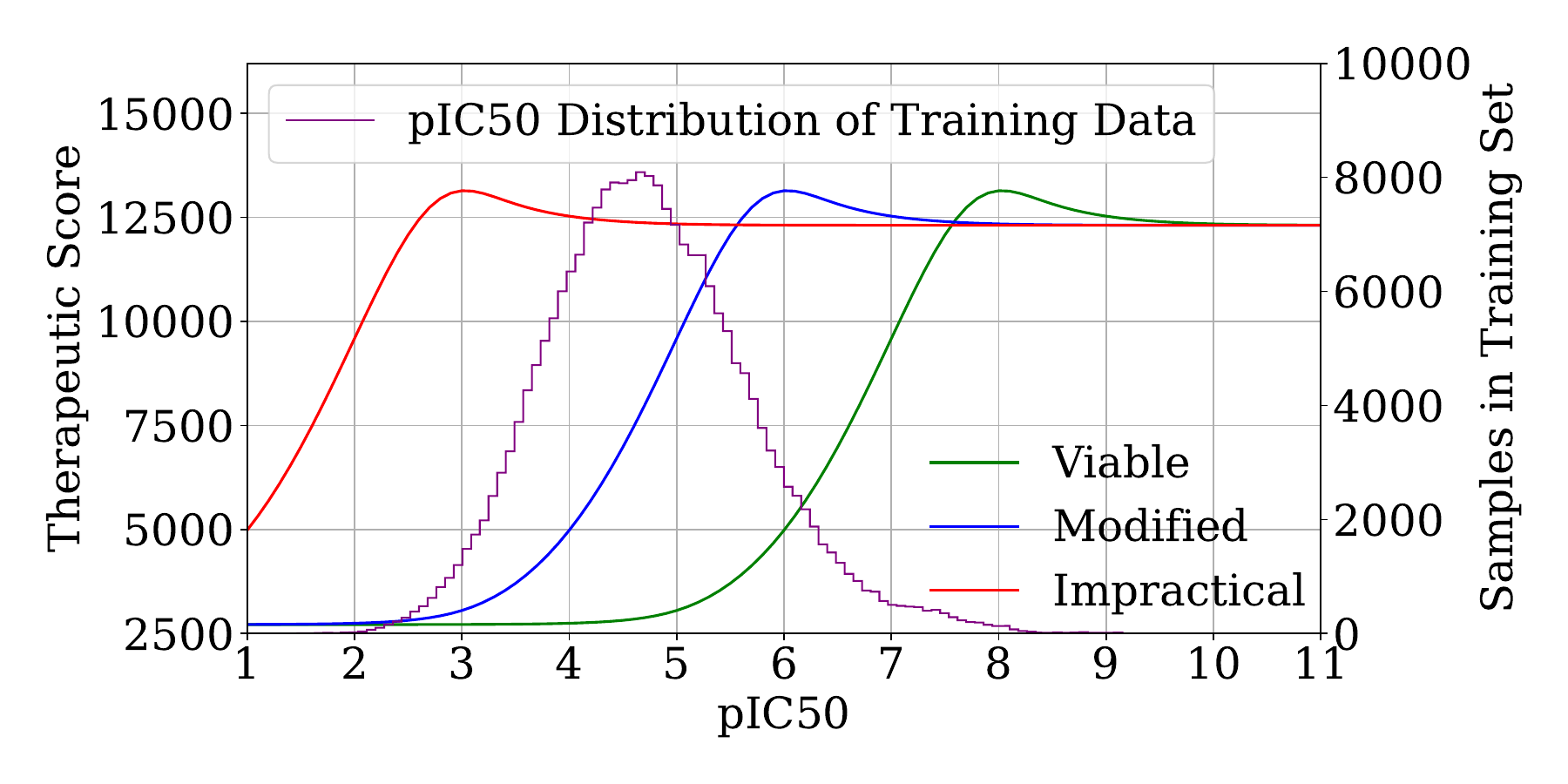}
    \vspace{-6mm}
    \caption{Relationship between therapeutic score and pIC50 according to the pathway guided mechanistic models.}
    \label{therap-pxc50-relation}
\end{wrapfigure}

The relationship between the therapeutic score and pIC50 value according to each version of the pathway model is shown in Fig. \ref{therap-pxc50-relation}. The relation between therapeutic score and pIC50 is approximately linear in the lower range of pIC50, reaches to a maximum therapeutic score at around pIC50=8 for physiologically viable pathway model, at around pIC50=6 for modified pathway model and at around pIC50=3 for physiologically impractical pathway model, and then stays at a fixed therapeutic score for successive pIC50 values. The three versions of the pathway models primarily differ in the position of the peak point on the relationship curve between therapeutic score and pIC50.
\subsubsection{PARP1 Inhibition Model}
PARP1 inhibition data was gathered from public and private databases including ChEMBL30 \cite{Mendez2019},  GoStar \cite{gostar}, and HiTS \cite{hits}. A basic curation pipeline was applied that includes compound standardization, relegation of all applicable values to pIC50's (negative log of the IC50 in molar), discarding outliers and averaging repeated measurements. Special attention was paid to assay descriptions for each measurement to ensure data comparability across different sources. The final dataset consisted of 9411 compounds. The data was partitioned into training, validation and test sets using an $80/10/10$ split. A comprehensive hyperparameter search was conducted for split type (scaffold or fingerprint), model type (RF, XGboosted RF, FCNN or graph convolutional NN), and feature set (ECFP4 fingerprints, MOE, Mordred or RDKit descriptors) in addition to parameters for each model type. The best performing model was selected based on the validation set performance of R-squared metrics. It is a graph convolutional NN fingerprint split model. Model performance is listed in Table \ref{table:reg}. The data curation, hyperparameter search, and model evaluation was conducted using the ATOM Modeling Pipeline version 1.4.2 \cite{Minnich2020}.
\begin{table}[htbp]
\centering
\small
  \caption{\ Performance of the PARP1 inhibition model}
  \label{table:reg}
  \begin{tabular*}{0.48\textwidth}{@{\extracolsep{\fill}}llll}
    \hline
    & R-squared & MAE & RMSE \\ 
    \hline
    Training Set & 0.363 & 0.756 & 0.937 \\
    Validation Set & 0.419 & 0.757 & 0.982 \\
    Test Set & 0.158 & 1.045 & 1.298 \\
    \hline
  \end{tabular*}
\end{table}
\subsubsection{Therapeutic Score Calculation Using Rule-based Pathway-Guided Mechanistic Model}
The pathway model utilized in this project was adjusted for PARP1 binding to DNA double strand breaks but derived from a previously published report defining cell death (via apoptosis) by two separate mediators, the transcription factor, p53, or a membrane associated kinase, AKT \cite{bogdal2013levels}.  
\begin{wrapfigure}{l}{0.55\textwidth}
\centering
    \includegraphics[width = 0.55\textwidth]{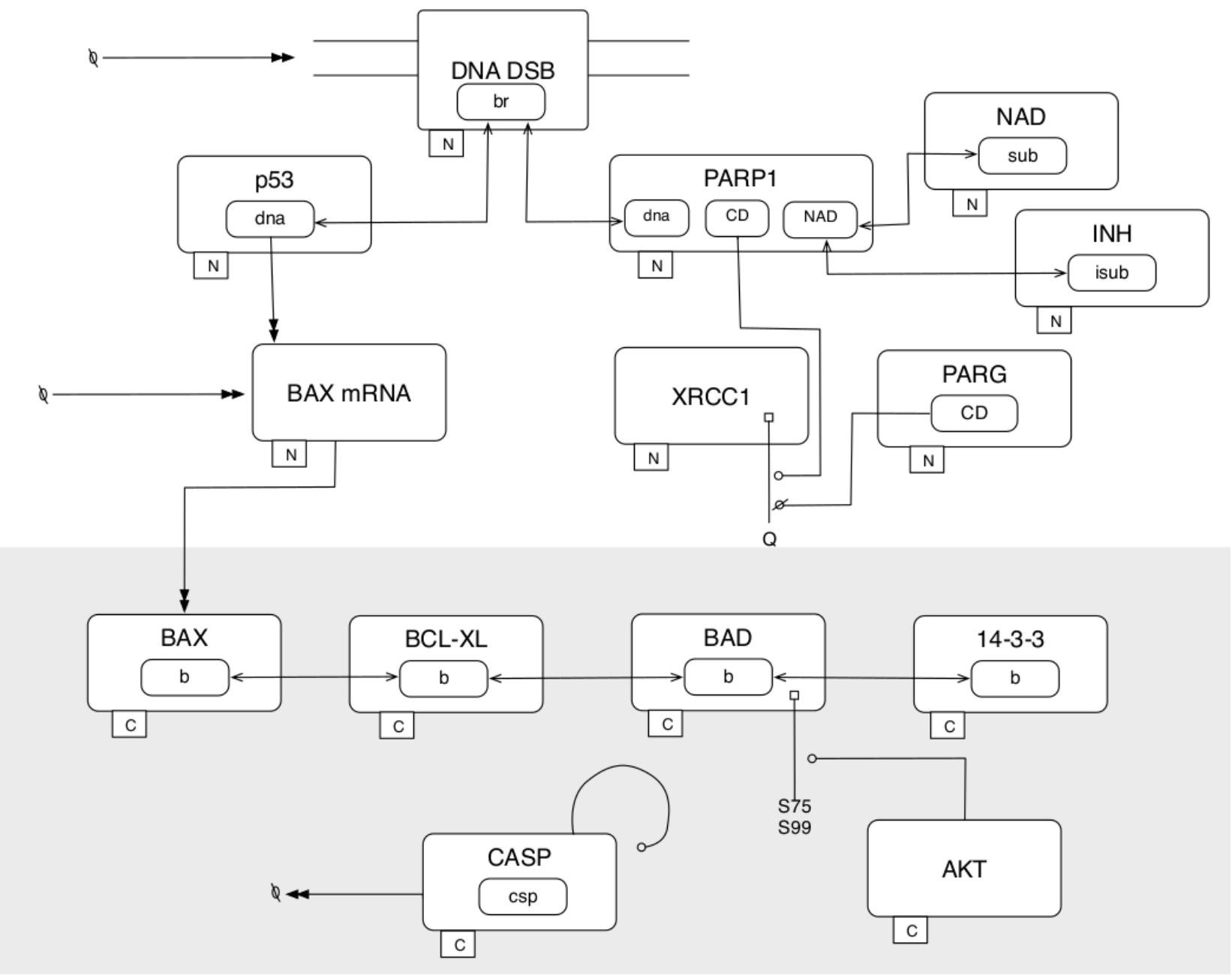}
    \caption{Extended contact map of pathway model. There are 14 molecules arranged for their location in the nucleus (white background) and the cytoplasm (shaded background). Nineteen reaction rules are depicted by the lines with and without arrows which further define binding and/or catalytic reactions.}
    \label{contact-map}
\end{wrapfigure}
In the updated model used here, the role of AKT is eliminated and only p53 action is considered through competition with PARP1 in the presence of DNA double stranded breaks. In the absence of an inhibitor, PARP1 functions to preserve cell viability in the presence of DNA double strand breaks. However, if PARP1 is bound to an inhibitor of sufficient strength, its function may become diminished, allowing p53 to trigger cell death via caspase activation. This caspase activated cell death is the therapeutic score. To reiterate, potent molecules with low IC50s in the model bind strongly to PARP1, leading to expected cell viability or death. And further, molecular abundance of activated caspase above 5000 was considered the trigger for apoptosis. The three versions of the pathway models are different according to their peaks. Fig. \ref{contact-map} presents an extended contact map of the reaction network included in the published pathway model that conforms to guidelines accommodating rule-based models.  Our methods utilized the reactions found in the white box and caspase activation (shaded box).
\section{Results \& Discussion}
\label{result}
\begin{figure*}[htbp]
\centering
    \includegraphics[width = 0.95\textwidth]{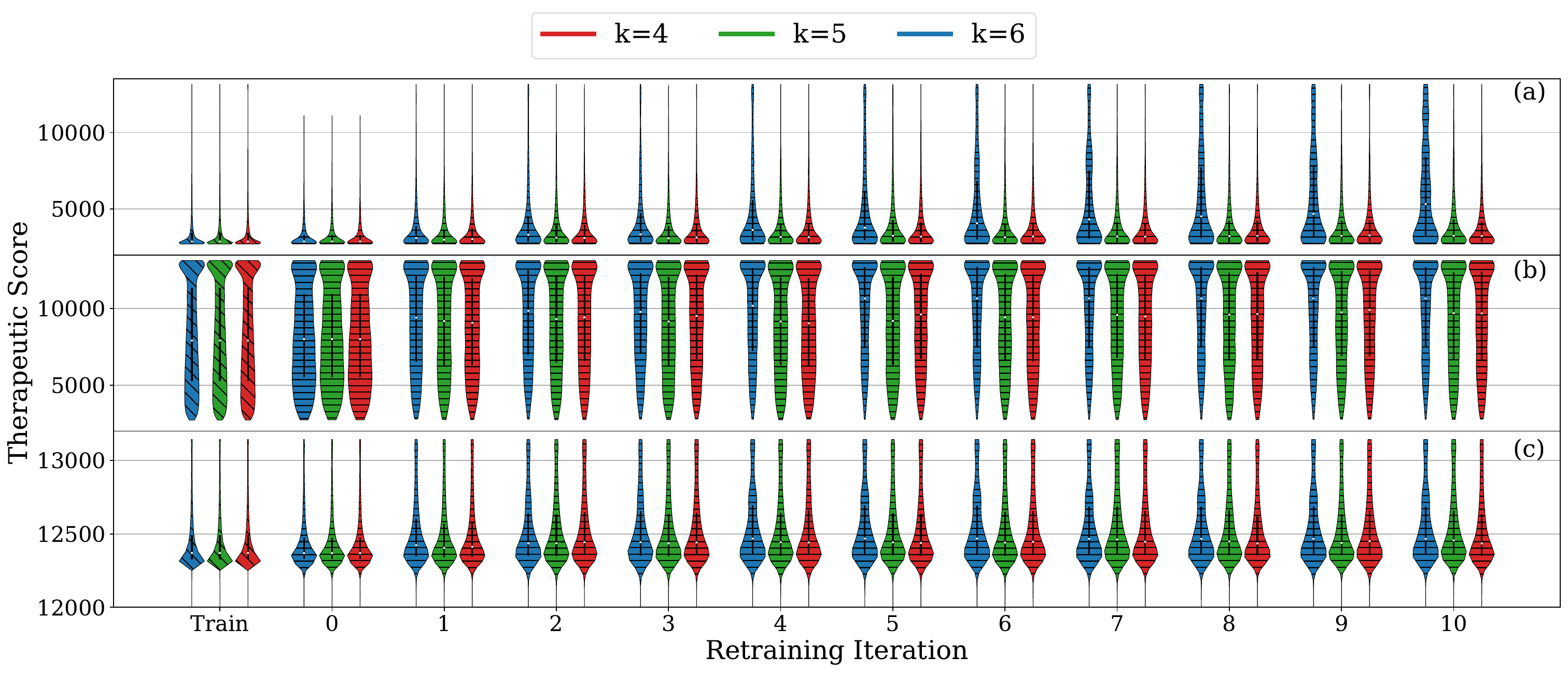}
    \caption{Distribution of therapeutic scores of generated molecules in consecutive retraining iterations for physiologically (a)viable, (b)modified, and (c)impractical pathway model.}
    \label{therap-dist}
\end{figure*}
\begin{figure*}
    \centering
    \includegraphics[width = 0.95\textwidth]{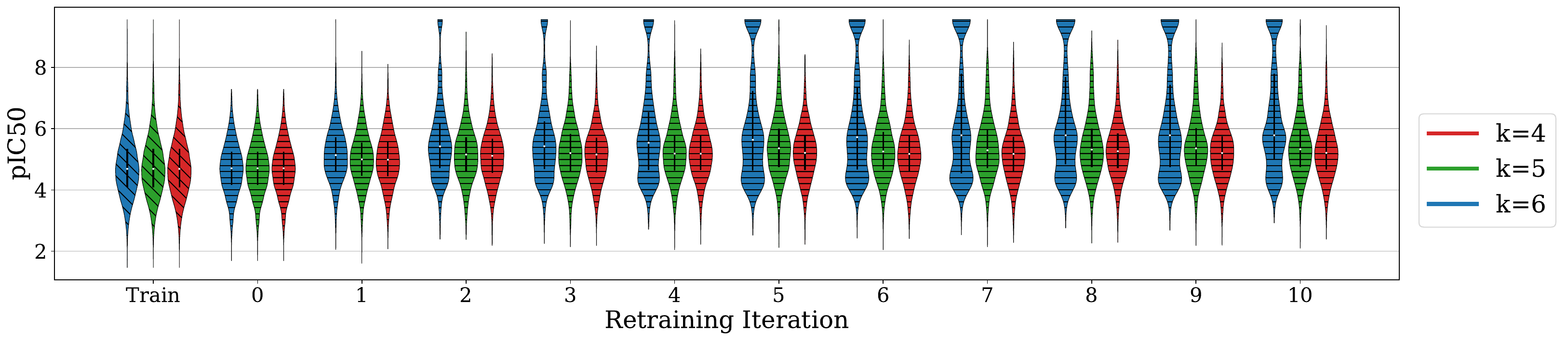}
    \caption{Distribution of pIC50 of generated molecules in consecutive retraining iterations.}
    \label{pIC50-dist}
\end{figure*}

Rather than embarking on the training of a JTVAE model from scratch, we employ a pretrained JTVAE model that has been previously trained in an unweighted fashion on the ZINC250K dataset \cite{zinc250k}. The initial training dataset consists of $228497$ small drug-like molecular structures. During the periodic weighted retraining and latent space optimization, the very first retraining iteration commences by utilizing the complete initial training dataset. In the subsequent retraining iterations, only a subset comprising $10\%$ of the original training datapoints is employed as the training set. This subset is complemented by the inclusion of optimized molecules that have been sampled through latent space Bayesian optimization. The optimized molecules that are sampled during one retraining iteration are employed to train the JTVAE model for all successive retraining iterations. We retrain for $10$ iterations with a retraining frequency of $50$, meaning at each retraining iteration we sample $50$ optimal points and append to the training dataset and carry on the retraining iterations. During every retraining iteration, we utilize Bayesian optimization to obtain a set of optimal data points sampled from the latent space. We employ a combination of $2000$ molecules with optimal scores and $8000$ molecules randomly selected from the training dataset to fit a Gaussian process model. It took about 12 hours to optimize the model through retraining for 10 iterations on a workstation with a single NVIDIA RTX3090 GPU. \newline
In this study, our primary emphasis is directed towards evaluating the impact of weighted retraining on the distribution of therapeutic score. In each retraining iteration, we take a random sampling of molecules from the latent space. This deliberate choice is motivated by our objective, which is to closely monitor and analyze the transformations occurring within the latent space as a result of the weighted retraining process. Following the completion of each retraining iteration, we proceed to randomly sample $5000$ molecules from the latent space. For the sake of fair evaluation, we use the same $5000$ randomly sampled latent space points at each retraining iteration to decode into molecular structures. \newline
To understand the effect of different weighting scales, we run the retraining experiments for different $k$ values. In this work, we observe the retraining effect for three different values of $k = 4, 5$ and $6$. Fig. \ref{therap-dist} shows the sample distributions of the generated molecules at consecutive retraining iterations for different $k$ values for all the pathway models. \newline
Fig. \ref{therap-dist} shows that $k=6$ pushes the model most to generate molecules with high therapeutic scores (typically $\geq 12000$) in consecutive retraining iterations for all the pathway models. The largest number of generated molecules with high therapeutic scores is observed in the physiologically impractical pathway model, followed by the modified pathway model, and the least in the physiologically viable pathway model. This phenomenon can be explained from the relation between therapeutic score and pIC50 for different pathway models, and the data distribution of the training set as shown in Fig. \ref{therap-pxc50-relation}. Intuitively, the greater the number of molecules with optimal therapeutic scores in the training set, the easier the optimization task becomes. For the physiologically impractical pathway model, the training set contains the largest number of molecules with optimal therapeutic scores, followed by the modified pathway model, and the least in the physiologically viable pathway model. Differences in training data distributions for different pathway models align the observations with our intuition. Additionally, the optimization process causes the greatest shift in sample distribution for the physiologically viable pathway model compared to the modified and physiologically impractical pathway models, due to the greater potential for distributional shift of the generated molecules in the viable pathway model.\newline
Figure \ref{pIC50-dist} illustrates the pIC50 distribution of the generated molecules when pIC50 is the optimization objective. As with the therapeutic score optimization, increasing $k$ drives the model to predominantly generate molecules with higher pIC50 values.

\section{Limitations of Proposed Approach}
\label{limitations}
There are three main aspects of our proposed methodology: periodic weighted retraining of JTVAE model, PARP inhibition model and rule-based mechanistic model. We discuss the limitations of these three different aspects of our proposed methodology.
\subsection{Limitations of Periodic Weighted Retraining of JTVAE}
\begin{figure*}
    \centering
    \includegraphics[width = 0.95\textwidth]{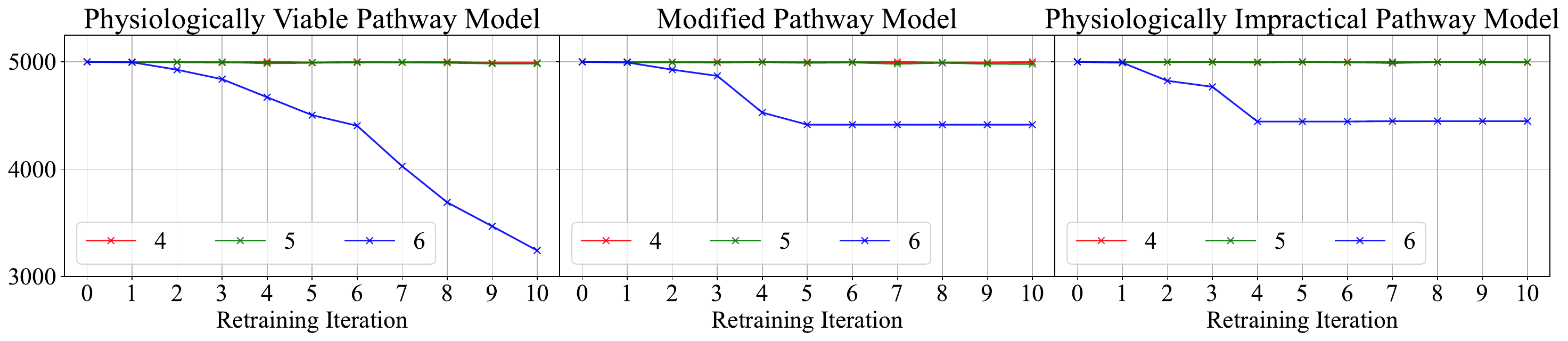}
    \caption{Number of generated unique molecules.}
    \label{fig:unique-mol}
\end{figure*}
Fig. \ref{fig:unique-mol} shows the number of generated unique molecules at each retraining iteration for different $k$ values and for all three pathway models. At each retraining iterations, the plot shows the number of generated unique molecules among the randomly generated 5,000 molecules from the latent space. With higher $k$ values, the number of uniquely generated molecules gets lower. From all the results shown, we see that $k=6$ experiments produce molecules with highest property values, but at the expense of reducing the latent space's capability of generating unique molecules. This means that weighting too much on high property molecules at the retraining phase may lead to the generation of high property molecules, but at the same time it converts the latent space in such a way that the generative model looses its capability of generating unique molecules. The decrease in unique molecules with each consecutive retraining iteration is the steepest for the physiologically viable pathway model. This reflects the fact that the training set contains the fewest molecules with optimal therapeutic scores for the physiologically viable pathway model.
\subsection{Limitations of PARP1 Inhibition Model}
The PARP1 inhibition model suffers from two main limitations. First, the majority of molecules with PARP1 inhibition data in the training set have medium to low potency values. The highest potency molecules are not structurally diverse and centered around a few select scaffolds like those of published clinical inhibitors. When optimizing toward high potency inhibitors of PARP1, the model may be biased toward these few scaffolds. To combat this, a fingerprint split model was used. Fingerprint splitting uses the Tanimoto similarity metric between Morgan fingerprints calculated from each molecule to ensure that the train, validation, and test sets are as structurally dissimilar as possible. During training, this should encourage the model to generalize its predictions beyond memorizing which scaffolds are the most potent. However, the test set performance metrics (Table \ref{table:reg}) show that the model does not necessarily generalize well to the new, structurally distinct molecules in the test set. These limitations can be especially detrimental for generative models where the goal is to propose novel and potent leads. If the model is biased or cannot generalize to new scaffolds well, it may steer the generative process toward the wrong area of chemical space.
\subsection{Limitations of Rule-based Pathway-guided Mechanistic Model}
There are two limitations of the pathway model utilized in this study. First, the pathway model focused on a small number of biochemical reactions in interest of limiting computational time. In particular, competition for DNA double strand breaks occurs between PARP1, the initiator DNA repair processes and p53, a known tumor suppressor, that is able to induce cell death in the advent of no repair. In fact, multiple repair processes exist as well as reaction networks leading to apoptosis. Thus, the absence of important details may alter systems dynamics such as timing of those events leading to different molecular abundances of activated caspases and calculated therapeutic scores. In particular, the pathway model used does not model the expected ultrasensitive activated caspase response found for other pathway models and living cells. Future studies may improve the usefulness of LOS if the pathway models better reflect biology. Another limitation of the pathway model, known for all mechanistic models, is parameter uncertainty. While every effort was made to utilized measured values reported in the literature, the pathway model only reflects one single cell, perhaps one of many heterogeneous ones in a tumor. Altering parameters or accounting for uncertainty in later computations may overcome this.
\section{Conclusions}
We present an optimization of a generative model combining deep generative models for molecular design with a rule-based, pathway-guided mechanistic model serving as the evaluation function for the generated molecular structures. This combined strategy aims to optimize the generation process of drug-like molecular structures. The results we've observed indicate that periodically retraining the generative model with adjusted weights gradually refines the model's parameters, resulting in the production of more optimal molecular structures. Additionally, our mechanistic model, guided by pathways, serves as the optimization objective function, negating the requirement for a dataset of molecular structures with known cancer therapeutic efficacy. 

\bibliographystyle{abbrvnat}
\bibliography{neurips_2024}

\end{document}